# Propositional and Relational Bayesian Networks Associated with Imprecise and Qualitative Probabilistic Assessments

**Fabio Gagliardi Cozman**[1]**, Cassio Polpo de Campos**[2]**, Jaime Shinsuke Ide**[1]**, José Carlos Ferreira da Rocha**[3]

[1]Escola Politécnica, Universidade de São Paulo, São Paulo, SP - Brazil
[2]Pontifícia Universidade Católica, São Paulo, SP - Brazil
[3]Universidade Estadual de Ponta Grossa, Ponta Grossa, PR - Brazil
fgcozman@usp.br, cassio@pucsp.br, jaime.ide@poli.usp.br, jrocha@uepg.br

**Abstract**

This paper investigates a representation language with flexibility inspired by probabilistic logic and compactness inspired by relational Bayesian networks. The goal is to handle propositional and first-order constructs together with precise, imprecise, indeterminate and qualitative probabilistic assessments. The paper shows how this can be achieved through the theory of credal networks. New exact and approximate inference algorithms based on multilinear programming and iterated/loopy propagation of interval probabilities are presented; their superior performance, compared to existing ones, is shown empirically.

## 1 INTRODUCTION

This paper focuses on a representation language that can accommodate propositional/first-order constructs and probabilistic assessments of various forms (precise, imprecise, indeterminate, and qualitative ones).

The oldest attempts to combine certain and uncertain beliefs go back to Boole's probabilistic logic [22]. Probabilistic logic was rediscovered by Nilsson [38], and has been extended to handle imprecise and qualitative beliefs, relations and quantifiers [3, 17, 23]. The price to pay for all this flexibility is computational complexity and, often, inferential vacuousness — inferences typically lead to probability intervals, and often these intervals are quite wide. A different strategy is adopted by relational Bayesian networks; here a strict set of rules produces a single probability measure and efficient inference [19, 27, 36, 41]. Perhaps the most important lesson from the Bayesian network literature is the importance of *structure*; that is, the importance of modular representations that admit fast inference algorithms.

It is natural to ask whether we can have structured representations that handle logical and probabilistic statements, some of which may be precise, imprecise, indeterminate or qualitative. This paper investigates representations with these characteristics, presenting basic theory, inference algorithms, and applications. Sections 2 and 3 discuss the basic aspects of our models. Section 4 then explores inferences, framing them in the context of *credal networks*. Sections 5 and 6 present several new algorithms for inference with credal networks, and experiments indicating that these algorithms surpass existing methods with respect to the size of handled networks. Section 7 discusses an example and Section 8 contains concluding remarks.

## 2 PROBABILISTIC LOGIC AND RELATIONAL BAYESIAN NETWORKS

Probabilistic logic associates probability values (which may be imprecise or indeterminate) to logical sentences [3, 17, 23, 38]. An example of application is *logic probabilistic programming*, where one can make statements such as

$$\text{KILLED}(x) \xleftarrow{[p,q]} \text{SHOT}(y,x) \wedge \text{LOADED}(y), \quad (1)$$

where $[p,q]$ is a probability interval — in fact several authors regard probabilistic imprecision as an important representation tool in itself [29, 32, 35]. Probabilistic logic offers a language for several classic artificial intelligence problems that deal with certain and uncertain beliefs; however, probabilistic logic faces two difficulties. First, it is computationally intractable in general. Second, inferences produce probability intervals, and it is easy to create situations where a few probability assessments lead to large probability intervals (call this *inferential vacuousness*).

"First-order" Bayesian networks also combine logic and probability but follow a different philosophy; they adopt as many assumptions as needed (independence relations, uniqueness of assessments) so as to yield efficient inference schemes. We now briefly review some relevant concepts related to such models. A *propositional Bayesian network* encodes a single probability measure over a fixed set of variables, using a directed acyclic graph, a node per variable, and associated conditional distributions [40]. First-order constructs can be added in several ways [19, 20, 27, 36, 41]; here we focus on *relational Bayesian networks* [27] as they are quite general and flexible. The idea is to have a vocabulary $S$ of relations, and to build a directed acyclic graph where each node is associated with a relation in $S$. Every relation $r$ in $S$ is then associated with a *probability formula* that indicates how to compute the probability of $r(v)$ for any $v$ in a domain $D$. The probability formula $F_r$ may depend on other relations, the parents of $r$ in the graph. When an inference is requested for some $r(v)$ (that is, a request for the probability of $r(v)$ conditional on some observed event), a propositional Bayesian network is generated; each node in this network represents the instantiation of a relation [28]. This "propositionalization" scheme requires methods for combination of probability formulas — for example, suppose that relation $r(v)$ has parent relations $s_i(u, v)$ that are valid for some $u$ satisfying relation $t(u)$; then the probability $p(r(v)|\{s_i(u, v) : (\forall u)t(u)\})$ must be specified. Relational Bayesian networks employ *combination functions*, denoted by $M\{F_1, \ldots, F_n | u; c(u, v)\}$, to combine $F_i$ for all $u$ that satisfy the equality constraint $c(u, v)$. Examples of such functions are Noisy-OR, Max, Min, and Mean.

Consider the following example.[1] If $v$ does not live in LA, then she sounds the alarm with probability 0.9 in case there is a burglary; if $v$ lives in LA, then she sounds the alarm depending on whether there is a burglary and whether there is an earthquake. This is expressed approximately as follows:
```
P(alarm(v)) = if (lives-in(v, LA))
                NoisyOR{
                  (burglary(v) ?  0.9:0.0),
                  (quake(LA) ?  0.2:0.0)};
              else
                (burglary(v) ?  0.9:0.0);
```
where $\text{NoisyOR}\{p_i\} = 1 - \prod_i(1 - p_i)$. This probability formula can be applied to any number of elements of $D$.

While probabilistic logic is *very loose*, relational Bayesian networks are *very strict*. In this paper we want to find some middle ground — structured and compact models that can deal with imprecision and indeterminacy. The next section proposes a methodology in this direction.

## 3 "RELATIONAL CREDAL" NETWORKS

Our proposal is to convey logical and probabilistic beliefs of various forms (precise, imprecise, qualitative) by associating directed acyclic graphs with *sets* of probabilities. The association of directed acyclic graphs with sets of probabilities is not new, as we discuss at the end of this section; our point is that this strategy offers an attractive "bridge" between the compactness of structured probabilistic models and the flexibility of probabilistic logic.

We assume that a directed acyclic graph captures a Markov condition; in this we simply follow the Bayesian-network philosophy. Thus a node[2] in the graph is independent of its nondescendants nonparents conditional on its parents. The graph topology is assumed known and captures the structure of the domain; for instance, the graph topology might come from a structured set of rules (1), or from direct causal information.

The best way to clarify this proposal is to discuss how imprecise, indeterminate and qualitative beliefs should be handled in a few concrete scenarios.

We start with a brief comment on uncertainty over probability values, as this type of uncertainty has been the object of extensive literature [30, 31, 47]. Several factors produce imprecision in probability values: there may be little subjective knowledge to obtain a precise value; there may be disagreement among experts in charge of a model; or probability values may be estimated through confidence intervals. The logical/probabilistic rule in Expression (1) is an example of probabilistic imprecision [29, 32, 35]; similar rules are found in relational databases, when measures of support and confidence are computed from data. It should be noted that induction of logic programs from finite data produces such interval-valued rules [5]. One might also consider more sophisticated ways to specify constraints on probability values, for example by belief functions or mass assignments [17]. In this regard, possibilistic databases offer an important example [4], as a possibility function can be readily interpreted as *upper* probabilities [16]. In all these situations, we obtain sets of distributions as representations for beliefs.

It is also important to recognize that *qualitative* statements of probabilistic strength yield sets of probabilities. Consider the *qualitative influences* that are employed in Boolean *qualitative probabilistic networks* [42, 48]: here a marked edge $Y \xrightarrow{+} X$ means that $P(x|y, z) \geq P(x|(\neg y), z)$, where $z$ denotes any instantiation of parents of $X$ except $Y$. This inequality typically defines a set of probabilities. The same is valid for mixtures of qualitative/quantitative assessments that have been proposed recently [42].

Consider now the possibility of imprecision and indeterminacy in combination functions. Take the most commonly used combination function, the NoisyOR function [20, 27, 36]. A NoisyOR function for Boolean variable $X$ and parents $\mathbf{Y} = \{Y_1, \ldots, Y_n\}$ depends on the *link* probabilities $P(x|y_i, \{(\neg y_j)\}_{j \neq i})$ — that is, the probability of $X$ given that $Y_i$ is true but all other parents are false. A difficulty is that these probabilities are not always available — most notably, the medical literature usually contains only *sensitivities* $P(x|y_i)$ and *specificities* $P((\neg x)|(\neg y_i))$ for each $Y_i$ [11]. One can try to translate sensitivities and specificities into link probabilities, but this translation is not unique [37]. The solution presented by Cooper in the NESTOR system [11] is to take sensitivities and specificities as *constraints* on the complete distribution $p(X|\mathbf{Y})$, a method that clearly produces sets of probabilities. The problem with this approach is inferential vacuousness, as there are too few constraints on the distribution $p(X|\mathbf{Y})$.

---

[1] Taken from the documentation of the Primula system, distributed at www.cs.auc.dk/~jaeger/Primula.

[2] A node/variable $X$ may refer to propositions or relations.

A better approach would be to examine exactly which assumptions behind the NoisyOR function are adequate to a problem, and adopt just these assumptions. For example, a NoisyOR function satisfies the following property:
**Cumulativity:** The more variables $Y_i$ are true, the larger is $P(x|\mathbf{Y})$.
This property, and a few others, may be used to characterize the NoisyOR function [14]. However, one might want to assume *only* a weakened form of this property in a particular situation; consider the following proposal:

$$\begin{aligned} p(X|\mathbf{Y}) &= \alpha \quad \text{if } Y_i = (\neg y_i) \text{ for all } i, \\ p(X|\mathbf{Y}) &= p_i \quad \text{if } \begin{cases} Y_i = y_i \text{ and} \\ Y_j = (\neg y_j) \quad \text{for } j \neq i, \end{cases} \\ p(X|\mathbf{Y}) &\geq \max\{p_i : Y_i = y_i\} \quad \text{otherwise.} \end{aligned} \quad (2)$$

Here we have $p(X|\mathbf{Y})$ always *larger than or equal to* the largest link $p_i$ among the "active" $Y_i$. Also we have a small "leak" probability $\alpha$ for the event $\{X = x\}$ conditional on $\{Y_i = (\neg y_i)\}_{i=1}^n$ (where $\alpha \leq \min_i p_i$). Call Expression (2) the *cumulative-synergy* model. The set-valued fuction (2) offers as much precision as possible given the assumptions.[3] A variant of this model is produced if sensitivities and specificities are given instead of link probabilities. For example, if a sensitivity $s_i = P(x|y_i)$ is given, then the equality $s_i = \sum_{\mathbf{Y} \setminus Y_i} p(X|\mathbf{Y}) P(\mathbf{Y} \setminus Y_i | y_i)$ must be satisfied — $p(X|\mathbf{Y})$ is given by the cumulative-synergy model with free parameters $p_i$, and $P(\mathbf{Y} \setminus Y_i | y_i)$ are free variables. Again, we are left with a set of probabilities over $X$ and $\mathbf{Y}$.

Qualitative relationships can also be used to constrain combination functions involving Boolean variables. For example, the *product/additive synergies* [48] define nonlinear constraints over the probabilities $P(X|\mathbf{Y})$. These qualitative assessments can be used in isolation or together with the constraints already discussed.

Hopefully at this point it is clear that many kinds of beliefs can be represented relational Bayesian networks associated with sets of probabilities. Suppose then that one builds such a model; now suppose an *inference* must be computed (that is, there is a request for probability bounds for some $r(v)$). We can start by creating an auxiliary propositional structure from the relational one, following the same procedures used in relational Bayesian networks [28]. This produces a directed acyclic graph and propositional variables associated with sets of probabilities — an object that has been extensively investigated and is known as a *credal network* [1, 7, 12, 18, 46]. A few relevant concepts are reviewed here.[4] A set of probability distributions is called a *credal set* [30]. A *conditional credal set* is a set of conditional distributions, obtained applying Bayes rule to each distribution in a credal set of joint distributions [47]. There are two kinds of conditional credal sets: if the distributions for $p(X|Y = y')$ and for $p(X|Y = y'')$ are unrelated, then the sets are *separately specified*; if these distributions are related, then the sets are *extensively specified* [43]. For example, the cumulative-synergy model is separately specified, while qualitative influences are extensively specified. Now consider a credal set containing joint distributions $p(X, Y|Z)$, and say that $X$ and $Y$ are *strongly* independent conditional on $Z$ if the vertices of this set factorize as $p(X|Z)p(Y|Z)$ (note that other concepts of independence for credal sets can be found in the literature, but strong independence seems to be the natural one in the present context) [12, 13]. The *strong extension* of a credal network is the largest joint credal set that satisfies a Markov condition: a variable is *strongly* independent of its nondescendants nonparents conditional on its parents [13].

Thus our proposal is to use "relational credal networks" to combine logical constructs and several forms of probabilistic assessments. Having identified the structure of interest, we now must look into inference procedures.

## 4 INFERENCES

We consider inference for a "propositionalized" credal network. Even though inference is a NP-complete problem for general polytree-shaped credal networks [43], a "pocket" of tractability is found in *Boolean* polytree-shaped credal networks, for which polynomial algorithms exist (Section 6 discusses this point in more detail).[5] We can thus state the following easy but notable result:

**Theorem 1** *If a relational credal network with Boolean variables is propositionalized into a network with polytree topology and separately specified credal sets, then inference is polynomial.*

In fact, a more general result can be stated: if we can divide a Boolean credal network in pieces, such that multiply connected pieces contain only singleton credal sets, and such that the various pieces form a polytree, then inference with the credal network is essentially as hard as inference with a Bayesian network of identical topology.

The following theorem clarifies the (yet open) complexity of inferences for multiply connected credal networks:

**Theorem 2** *Inference with the strong extension of a credal network is $NP^{PP}$-complete.*

---

[3]An attractive property of the cumulative-synergy model is that it admits "internal" factorizations as the NoisyOR function; due to lack of space, such computational properties are omitted.

[4]Several tutorials can be found at the Society for Imprecise Probability Theory and Applications, www.sipta.org.

[5]Another pocket of tractability is represented by purely qualitative networks [42, 48]; however in this paper we focus on models that can combine qualitative and numeric assessments.

*Sketch of proof.* The proof follows the same arguments in Park's theorem for the MAP problem [39]. Membership in NP$^{PP}$ is immediate. Hardness is shown by reduction of E-MAJSAT; Park's theorem shows a reduction to a MAP problem that is equivalent to inference in a credal network using the Cano-Cano-Moral transform [12]. QED

In short, inferences are in NP and NP$^{PP}$ completeness classes — exactly the classes that contain MPE and MAP problems for Bayesian networks. Thus we are within the confines of currently used probabilistic inference.

## 5　INFERENCE ALGORITHMS BASED ON MULTILINEAR PROGRAMMING

Consider the computation of a tight upper bound for $P(x)$ (the *upper* probability of $\{X = x\}$); this is obtained as

$$\max \sum_{X_1,\ldots,X_n \setminus X} \prod_{i=1}^{n} p(X_i|\text{pa}(X_i)), \quad (3)$$

subject to constraints on the distributions $p(X_i|\text{pa}(X_i))$. Exact inference algorithms for credal networks follow either enumeration or search methods to find the maximizing distributions [9, 12, 43, 44]. Despite intense effort, relatively "small" inferences have been processed exactly so far (about 15 nodes for polytrees, about 8 nodes for multiply connected networks with ternary variables). Several authors have suggested the direct use of nonlinear optimization for inference [1, 12, 18], but no algorithm has yet been formulated using this approach. The objective of this section is to investigate and implement the idea.

For the constraints discussed in Section 3, Problem (3) is a multilinear program on free variables $p(X_i|\text{pa}(X_i))$. If an upper bound for a *conditional* probability is requested, then a fractional multilinear program must be solved. As a fractional multilinear program can be solved by a sequence of multilinear programs [2], we only discuss Problem (3).

### 5.1　INFERENCE AS A MP PROBLEM

Problem (3) has the unpleasant property that the objective function contains an exponential number of terms (exponential on the size of the given credal network). This difficulty can be avoided by introducing new artificial variables that stand for summations in Expression (3). To illustrate this procedure, consider a simple network $A \to B \to C \to D \to E$. Assume all variables in the network are ternary. Computation of the upper probability for $\{E = e_0\}$ using Expression (3) leads to $\max \sum_{h,i,j,k} p(e_0|d_h)\, p(d_h|c_i)\, p(c_i|b_j)\, p(b_j|a_k)\, p(a_k)$, a multilinear function with 81 nonlinear terms of degree four. We can transform this expression by introducing new variables so as to keep the degree at most 2. We obtain just 30 nonlinear terms in $\max \sum_i p(e_0|d_i)\, p(d_i)$ subject to $p(d_k) = \sum_j p(d_k|c_j)\, p(c_j)$, $p(c_k) = \sum_j p(c_k|b_j)\, p(b_j)$, $p(b_k) = \sum_j p(b_k|a_j)\, p(a_j)$ (for $k = 0, 1, 2$), plus the linear constraints.

This "decomposition" of Expression (3) can be automated using the variable elimination algorithm for Bayesian network inference. Run this algorithm and define new artificial variables for each value of the intermediate functions generated during variable elimination. The objective function becomes a summation of a few of those artificial variables; each new artificial variable corresponds to a multilinear expression representing relationships between neighbour nodes in the elimination tree. The number of functions in the resulting MP problem is linear on the number of parameters of the credal network. The decomposition is quite fast and essentially takes the cost of a single Bayesian network inference.[6] Details can be found elsewhere [6].

We now consider the solution of the resulting MP problems. We should stress that an advantage of such a "direct" optimization scheme is that constraints on probabilities can be nonlinear (for example, the qualitative constraints discussed in Section 3) and credal sets need not be separately specified. Existing algorithms for inference with credal networks typically cannot handle such situations. We thus consider a multilinear programming (MP) problem formulated as $\max f_0(\theta)$, where $\theta$ contains probability values that belong to a box in $\mathbb{R}^m$ and satisfies constraints $\sum_{t \in T_r} \alpha_{rt} [\prod_{j \in J_{rt}} \theta_j] \geq \beta_r$ for $r = 1, \cdots, R$; $T_r$ is an index set defining terms of these constraints, $\alpha_{rt}$ is the real coefficient for the $t$th term, and $J_{rt}$ indicates the set of variables in the $t$th term. Define $T = \cup_r T_r$. MP problems are nonconvex and no known transformation can convexify them; several solution methods have been proposed in the literature [2, 21, 24, 33, 45].

The properties of Sherali and Tuncbilek's Reformulation-Linearization (RL) method [45] make it particularly appropriate to solve Expression (3). The RL method substitutes each product of variables $\prod_{j \in J_{rt}} \theta_j$ by a new artificial variable $\vartheta_{J_{rt}}$ for all terms $t \in T$, thus obtaining a linear program. The solution of each linear problem gives an upper bound to the solution of the MP problem. The method iterates over the variables by branching over their ranges whenever necessary, until each $\vartheta_{J_{rt}}$ is close enough to $\prod_{j \in J_{rt}} \theta_j$. To guarantee convergence, some additional "artificial" functions have to be included in the linear subproblems; these variables correspond to multiplications of the original constraints (provided that the degree of the new functions do not exceed the maxdegree $\delta = \max_{r,t} |J_{rt}|$, because this would increase the complexity of the problem) [45]. In our implementation we only construct new artificial functions when the terms they refer to are already present in the MP problem. To choose a variable to branch over, the method finds the greatest difference between the artificial variables and the products they represent, and di-vides the range of that variable. Every time a linear solution is feasible for the MP problem, the method verifies whether it is the best solution known so far. Thus the branching steps are obtained by "cuts" on the feasible region.[7] Finally, we note that branching in the RL method can benefit from knowledge of local maxima [45]; in our implementation we use the search algorithm by Rocha et al [44] to produce local maxima.

---

[6]Andersen and Hooker [1] proposed a similar decomposition, without specifying an algorithm for it; they also suggested that a direct MP solution would be necessary, but did not present any algorithm for it.

Table 1: Test sets (each with 10 networks), with average sizes of MP problems and their corresponding linearized versions, average sizes of intervals containing inference (always an upper probability), and the average number of branches examined by the RL method. A few networks in the fourth row, and half the networks in the fifth row could not be solved.

| Network Topology | Nodes | Vertex by credal set | MP vars. | MP funcs. | Linear. vars. | Linear. funcs. | A/R++ error | RL error | Branched nodes |
|---|---|---|---|---|---|---|---|---|---|
| dense Boolean | 10 | 2 | 105 | 172 | 665 | 3996 | 2.8684% | 0.0484% | 301 |
| Alarm Boolean | 37 | 2 | 363 | 576 | 1395 | 6876 | 5.5706% | 1.076% | 765 |
| dense ternary | 10 | 3 | 412 | 576 | 5920 | 40181 | 10.4304% | 0.3290% | 1 |
| Alarm ternary | 37 | 3 | 1657 | 2214 | 13780 | 70612 | 22.3293% | 2.5954% | 3 |
| dense quaternary | 10 | 4 | 1145 | 1474 | 30073 | 213376 | 13.4146% | 0.6071% | 1 |

### 5.2 THE A/R++ ALGORITHM

The performance of the RL method is greatly enhanced if ranges for free variables $\theta$ and $\vartheta$ are known [45]. One way to obtain approximate ranges for polytree-shaped networks is to run the A/R+ algorithm [43]. The A/R+ algorithm (and Tessem's original A/R algorithm [46]) work by producing local approximations for the messages sent during inference in polytrees. A node $X$ receives approximate interval probabilities from its parents and children, and sends approximate interval probabilities to its parents and children; these approximate interval probabilities are quickly computed and transmitted. Hence we can use the A/R+ algorithm to improve the RL method.

Now, we can also use the RL method to improve the A/R+ algorithm. Take a node $X$ and consider that $X$ must send a message $\lambda_X(Y)$ to its parent $Y$, by combining messages received from $X$'s other parents and children. The A/R+ algorithm sends upper and lower bounds for $\lambda_X(Y)$, and these upper and lower bounds can be *easily produced by multilinear programming* — they are actually local versions of Expression (3) [46]. In fact, bounds on the probability of any event defined by $X$, conditional on $Y$, can also be produced by multilinear programming. Thus we obtain the following algorithm, which we call *A/R++*: follow the same steps of the A/R+ algorithm, but compute probability bounds for several events using multilinear programming, and send these bounds as messages. The probability intervals computed by A/R++ are always more precise than or equal to the intervals of the A/R+. Note that the additional bounds that are passed amongst variables are linear

---

[7]In our context, these cuts are cuts on the credal sets in the credal network; this branching strategy is different from the branching method proposed by Rocha et al [44], where each branch corresponds exactly to a vertex of a credal set.

constraints that can be easily handled by the RL method.

So far we have discussed the A/R++ algorithm as a method for polytree-shaped networks; the algorithm can be readily extended to multiply connected networks, by considering messages in the variable elimination algorithm instead of messages directly amongst nodes. Technical details can be found elsewhere [6].

### 5.3 EXPERIMENTS

We have conducted experiments on five sets of networks, to illustrate the behaviour of inference with A/R++ and our RL-based method. Results are shown in Table 1. Each test set was composed of 10 randomly generated multi-connected credal networks (generated with BNGenerator [25]). Experiments refer to computation of upper probabilities without evidence; results refer to the most challenging inferences in each network. Table 1 indicates the topology of the test networks and the size of multilinear and linearized programs. The size of linearized programs grows substantially with the number of vertices on credal sets and the number of variables in the credal network; for the larger networks, only a few branches in the RL algorithm are possible. The results also allows us to compare the quality of results produced by the A/R++ algorithm and the RL method (note that the RL method uses the A/R++ to produce ranges of variables). Experiments were performed in a Pentium IV 1.7GHz, using CPLEX as linear solver, and with a time-limit of ten minutes for the first three test sets and one hour for the other two test sets.

## 6   ALGORITHMS BASED ON ITERATED AND LOOPY PROPAGATION

Approximate inference seems to be a natural solution for large credal networks [8, 9, 10, 44, 46]. In this section we propose two new approximate inference algorithms that are geared towards Boolean networks, given the importance of such networks in relational settings; the discussion is brief and technical details can be found elsewhere [26].

As Boolean polytrees have polynomial inference (Theorem 1), we focus on Boolean multiply connected networks. Our new algorithms rely on the 2U algorithm — the first poly-nomial inference method for Boolean polytrees [18]. The 2U algorithm slightly modifies the structure of messages used in Pearl's belief propagation algorithm [40]. In the 2U algorithm, each node $X$ computes values $P(X = x|E)$ by combination of interval functions $\pi(X)$ and $\Lambda(X)$ — these interval functions are produced by processing several "messages" received by $X$. A complete account of the 2U algorithm can be found in the original paper [18].

### 6.1 ITERATED PARTIAL EVALUATION (IPE)

Draper and Hank's Localized Partial Evaluation (LPE) algorithm produces approximate inferences by "cutting" parts of a network and running interval-based inferences in a selected sub-network [15] (in Boolean networks, we can use the 2U algorithm for the interval-based inferences). We propose the following algorithm:
1) Select a conditioning cutset [40] for the credal network.
2) "Cut" the Boolean network (using the cutset and the LPE operations) so that the resulting network is a polytree.
3) Now run LPE on the polytree, using 2U as the inference engine. In polynomial time we obtain an approximate probability interval for any node in the credal network.
4) Select a different cutset, and return to Step 2, for a given number of iterations.
5) At the end, return the intersection of all approximate probability intervals generated in the process.
We have:

**Theorem 3** *The probability interval produced by the IPE algorithm contains the exact interval requested by the inference.*

*Sketch of proof.* Each run of the LPE+2U algorithm contains the exact interval, because the LPE algorithm processes all vertices of credal sets in the network; the intersection of all approximate intervals contains the exact interval. QED

We have implemented the IPE algorithm and run experiments in the network topologies employed by Murphy et al [34] to test loopy propagation: the Pyramid and the Alarm networks. The Pyramid network is a multilayered graph associated with Boolean variables and local connections among layers. The Alarm network is a classic model used in medical diagnostic; we set all variables to Boolean values, so as to run the IPE algorithm. For both networks, we generated several realizations of random, uniformly distributed conditional probability tables [25]. Results can be viewed in Figure 1; most inferences are quite accurate, with mean square error (MSE) of 5% for Pyramid and 7.2% for the "Boolean" Alarm.

### 6.2 LOOPY 2U (L2U)

A popular algorithm for approximate inference in Bayesian networks is loopy propagation [34]. Here we propose a "loopy" variant of the 2U algorithm for multiply connected Boolean credal networks. First, a sequence of nodes **S** is randomly chosen, such that every node that is relevant to the inference is in **S**. Initialization of variables and messages follow the same steps used in the 2U algorithm. Then the nodes are repeatedly updated following the sequence **S**. Iterations are indexed by $i$, which starts at 1, and updates are repeated until convergence of probabilities is observed or until a maximum number of iterations is reached. Each node is updated in several steps:
1) Update $\underline{\pi}^{(i+1)}(x)$ and $\overline{\pi}^{(i+1)}(x)$ using $\underline{P}(X = x|Z_1, \ldots, Z_m)$, $\overline{P}(X = x|Z_1, \ldots, Z_m)$ and messages $\pi_X(Z_i)$ from the $m$ parents $Z_i$ of $X$.
2) Compute $\underline{\Lambda}_X^{(i+1)}$ and $\overline{\Lambda}^{(i+1)X}$ using messages $\underline{\Lambda}_{Y_j}^{(i)}(X)$.
3) Compute interval messages to be sent to the children and parents of $X$, $\pi_{Y_j}^{(i+1)}(x)$ and $\Lambda_X^{(i+1)}(Z_i)$, using the values computed at $(i)$.
The whole algorithm follows the loopy propagation scheme, but instead of messages from Pearl's propagation algorithm, here we use interval messages from the 2U algorithm. It should be noted that computation of interval functions require effort $\mathcal{O}(2^{2m})$, where $m$ indicates the number of parents of $X$ [18]. Consequently, the overall worst-case complexity of the L2U algorithm is $\mathcal{O}(k 2^{2m^*})$, where $k$ is the maximum number of iterations, and $m^*$ is the maximum number of parents in the network.

We have implemented L2U and run tests in the same networks used to test the IPE algorithm. The L2U algorithm converged after 4 iterations in the Pyramid network, and after 9 iterations in the "Boolean" version of the Alarm network. The mean square error (MSE) of several approximate inferences was only 1.3% for both networks; these results can be viewed in Figure 1. It should be noted that L2U generally produces approximate inferences quite quickly: inferences for the "Boolean" Alarm network were produced in less than one second in a Pentium computer.

## 7 EXAMPLE: THE *HOLMES* NETWORK

It is perhaps useful to show a complete example, however simple, of inference with a relational credal network. The purpose here is to evaluate the extent that inferential vacuousness can be a difficulty; computational aspects of inference algorithms have been discussed in previous sections.

Take then the *Holmes* example [36, 27], as described in the Primula system (Section 2). A person does or does not live in LA; the person's house is burglarized or not; there may be an earthquake in LA; and the person will or not sound the alarm, depending on the burglary and on the earthquake. The critical probability formula for this problem was given at the end of Section 2.

Consider a modification of the original *Holmes* network, where the prior probabilities are imprecisely known: $P(\texttt{burglary}(v)) \in [0.001, 0.01]$, $P(\texttt{earthquake}(LA)) \in [0.01, 0.1]$ and $P(\texttt{lives-in}(v, LA)) \in [0.05, 0.15]$. Suppose also that the NoisyOR function is replaced by a cumulative-synergy model with identical link probabilities. The evaluation of the leak probability for the cumulative-synergy model is a difficult matter, as leaks simply stand for unexplored territory; assume the leak probability to be imprecise, in the interval $[0.0, 0.1]$.

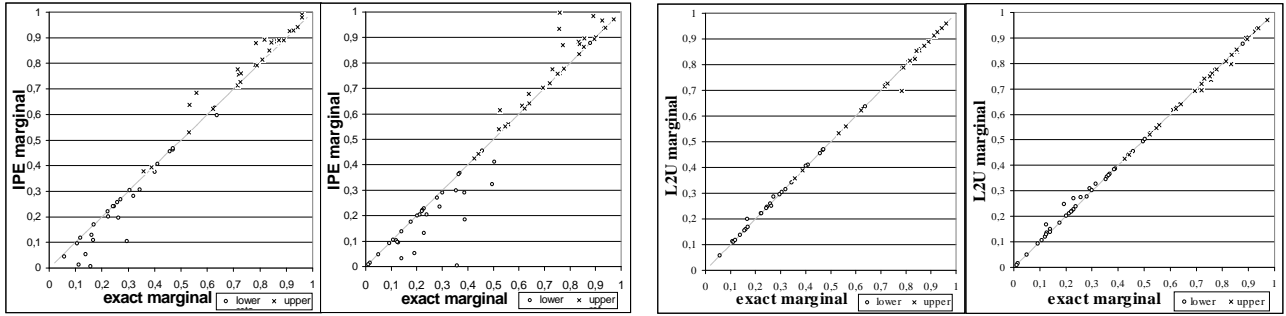

Figure 1: Experiments with Pyramid network (20 variables, no evidence) and "Boolean" Alarm (37 variables, no evidence). From left to right: Pyramid network with IPE (100 iterations); "Boolean" Alarm network with IPE (100 iterations); Pyramid network with L2U (4 iterations); "Boolean" Alarm network with L2U (9 iterations).

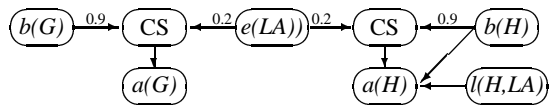

Figure 2: A "propositionalized" instance of the *Holmes* network, where: CS $\to$ cumulative-synergy (with link probabilities over edges); $a \to$ alarm, $b \to$ burglary, $e \to$ earthquake, $l \to$ lives-in.

For a domain containing $G$, $H$, $LA$, such that lives-in$(G, LA)$ is true, the credal network in Figure 2 is obtained. Consider a few inferences. The (unconditional) probability for alarm$(H)$ is in the interval [0.0001,0.0253]; if there is an earthquake in LA, the conditional probability is in the interval [0.0108,0.0388]. For $G$, we obtain $P(\text{alarm}(G)) \in [0.0029, 0.1179]$ and $P(\text{alarm}(G)|\text{earthquake}(LA)) \in [0.2007, 0.2080]$. The important point here is that inferences produce rather small intervals — even though only a few assessments are stated, the presence of structural assumptions on the domain greatly constrains the probabilities, obviating difficulties with inferential vacuousness.

## 8 CONCLUSION

The contributions of this paper can be divided in two groups.

First, we have proposed relational credal networks as a suitable language for certain and uncertain beliefs. Relational credal networks can handle several kinds of logical and probabilistic assessments, as discussed in Section 3. Even though the goals of probabilistic logic and relational Bayesian networks are closely related, our proposal seems to be the first explicit attempt to connect the two fields. As a suggestive example of application, consider the construction of a system for evaluation of monetary policy; it would be advisable to take the following piece into account:

> ...uncertainty is not just a pervasive feature of the monetary policy landscape; it is the defining characteristic of that landscape. The term "uncertainty" is meant here to encompass both "Knightian uncertainty," in which the probability distribution of outcomes is unknown, and "risk," in which uncertainty of outcomes is delimited by a known probability distribution (A. Greenspan, January 3, 2004)

Second, we have contributed with new theory and several new algorithms for inference in propositional credal networks (and to "propositionalized" relational credal networks). Inference with propositionalized models is polynomial for Boolean polytree-shaped networks (Theorem 1), and equivalent to standard MAP problems in general networks (Theorem 2). The RL-based method and the A/R++ algorithms (Section 5) are the first direct application of multilinear programming to inference; we have presented tests showing their effectiveness. We have also presented new iterative and loopy approximate algorithms that produce excellent results with short execution times (Section 6). The L2U algorithm is particularly promising, even though a solid convergence analysis is missing at this point. Taken together, our experiments indicate that existing medium-size topologies such as the Alarm network, can now be processed exactly, and much larger networks can be processed approximately. Clearly several challenges are yet to be overcome, but we hope to have demonstrated the feasibility of inference with our proposed models.

### Acknowledgements

This work has received generous support from HP Labs and HP Brazil. The work has also been supported by CNPq, CAPES and FAPESP.